\documentclass[11pt]{article}

\usepackage{acl}

\usepackage{times}
\usepackage{latexsym}
\usepackage[T1]{fontenc}    
\usepackage[utf8]{inputenc} 
\usepackage{microtype}      
\usepackage{inconsolata}   
\usepackage[detect-all]{siunitx}

\usepackage{amsmath}
\usepackage{pifont}
\usepackage{xcolor}

\usepackage{graphicx}
\usepackage{url}

\usepackage{booktabs}
\usepackage{multirow}
\usepackage{array}
\usepackage{threeparttable}
\usepackage{makecell}
\usepackage{calc}

\usepackage{enumitem}

\usepackage{algorithm}
\usepackage{algorithmicx}
\usepackage{algpseudocode}

\usepackage{xcolor}
\usepackage{tcolorbox}
\definecolor{promptbg}{HTML}{D0D8FF}
\definecolor{prompttitle}{HTML}{6882FE}

\newtcolorbox{promptbox}[1][]{%
  colback=promptbg,      
  colframe=prompttitle,   
  coltitle=white,         
  title=#1,               
  fonttitle=\bfseries,    
  fontupper=\sffamily\small, 
  boxrule=0.5pt,          
  arc=2mm,                
  top=2mm, bottom=2mm, left=2mm, right=2mm
}

\usepackage{hyperref}

\title{Text-to-SQL as Dual-State Reasoning: Integrating Adaptive Context and Progressive Generation}

\author{
  Zhifeng Hao$^{1,2}$ \quad
  Qibin Song$^1$ \quad
  Ruichu Cai$^{1,3}$ \quad
  Boyan Xu$^{1}$\thanks{\ \ Corresponding author, \url{hpakyim@gmail.com}} \\
  $^1$School of Computer Science, Guangdong University of Technology \\
  $^2$College of Science, Shantou University \\
  $^3$Peng Cheng Laboratory \\
  \texttt{haozhifeng@stu.edu.cn} \\
  \texttt{\{gdutsongqibin, cairuichu, hpakyim\}@gmail.com}
}

\begin{document}
\maketitle

\begin{abstract}
Recent divide-and-conquer reasoning approaches, particularly those based on Chain-of-Thought (CoT), have substantially improved the Text-to-SQL capabilities of Large Language Models (LLMs). However, when applied to complex enterprise databases, such methods struggle to maintain coherent reasoning due to limited context capacity, unreliable schema linking, and weak grounding in database semantics. To overcome these issues, we introduce DSR-SQL, a \textbf{D}ual-\textbf{S}tate \textbf{R}easoning framework that models Text-to-SQL as an interaction between an adaptive context state and a progressive generation state. The first constructs a compact, semantically faithful environment by refining large schemas and selecting relevant structures, while the second formalizes SQL synthesis as feedback-guided state transitions, enabling the model to self-correct and align with user intent. Without any post-training or in-context examples, DSR-SQL achieves competitive performance, reaching 35.28\% execution accuracy on Spider 2.0-Snow and 68.32\% on BIRD development set. Our implementation will be open-sourced at: \url{https://github.com/DMIRLAB-Group/DSR-SQL}.
\end{abstract}

\section{Introduction}
Text-to-SQL enables non-technical users to access and analyze structured data through natural language, bridging the gap between human intent and relational databases \citep{qin2022surveytexttosqlparsingconcepts, sui2023reboostlargelanguagemodelbased, chen2025texttosqlenterprisedataanalytics, liu2025survey}. 
In enterprise environments, this capability forms the core of many business intelligence systems—powering ad-hoc querying, dashboard generation, and multi-table analytics that support operational and strategic decision-making.

Existing Text-to-SQL benchmarks such as Spider capture only moderate schema complexity and thus fail to reflect the realities of enterprise-scale databases, which often span hundreds of interdependent tables and evolve continuously over time. 
More recent corpora like Spider~2.0 \citep{lei2024spider2} and BIRD \citep{li2023llm} reveal the key bottleneck in such settings: as schema size and domain heterogeneity grow, LLMs struggle to maintain accurate \emph{query–schema alignment} within a limited context window. 
Without sufficient grounding in database semantics, a single-pass generation process tends to produce SQL that is syntactically correct but semantically misaligned with user intent. 
Overcoming this misalignment in a zero-shot setting—without labeled data or task-specific tuning—remains an open challenge for both research and practical deployment \citep{Chen_Guo_Wang_Qiu_Qi_Wang_Li_2021, liu2023comprehensiveevaluationchatgptszeroshot, dong2023c3zeroshottexttosqlchatgpt, li2025alphasqlzeroshottexttosqlusing}.

\begin{figure*}[t!]
\centering
\includegraphics[width=1\linewidth]{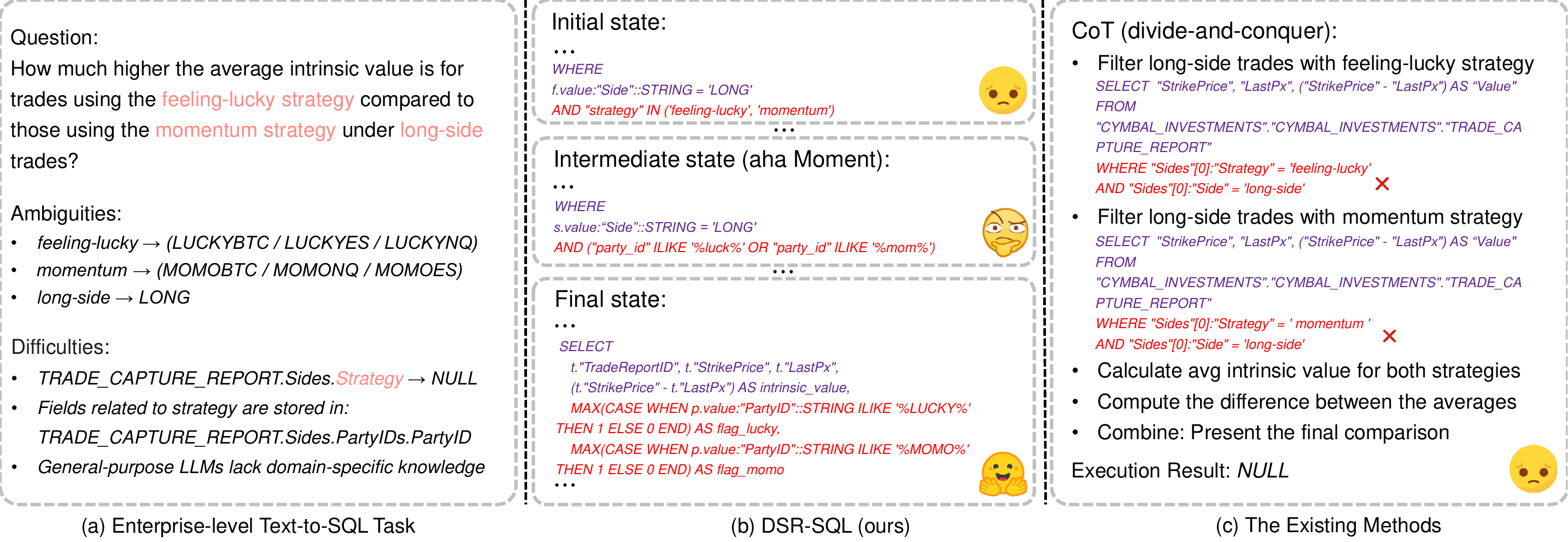}
\caption{
A motivating example illustrating the need for dual-state reasoning. 
Panel~(a) shows an enterprise-level Text-to-SQL query containing domain-specific terminology from quantitative trading, such as \textit{feeling-lucky strategy}, \textit{momentum strategy}, and \textit{long-side trades}. 
The database, however, stores these entities in entirely different formats. 
Panel~(c) demonstrates the failure of static, CoT-based divide-and-conquer reasoning, which splits the problem linguistically but remains unaware of the true schema, resulting in execution failure. 
In contrast, panel~(b) shows \textbf{DSR-SQL}, which leverages execution feedback to revise intermediate states, probe the database when mismatches occur, and ultimately reach an “aha moment” that correctly aligns user intent with database structure.
}
\label{fig:FIG1_toy}
\end{figure*}

Prior work has explored two main strategies to handle large and complex schemas. 
The first focuses on \emph{schema compression and retrieval}, where methods such as retrieval-augmented generation or locality-sensitive hashing prune the schema to fit within model context limits \citep{talaei2024chesscontextualharnessingefficient, liu2025xiyansqlnovelmultigeneratorframework}. 
While these techniques improve efficiency, they introduce irreversible information loss: once a relevant table or column is omitted, the model can no longer recover the missing linkage during generation. 
A complementary line of research attempts to \emph{decompose} complex questions into sub-problems through static CoT prompting \citep{pourreza2023dinsqldecomposedincontextlearning,xie2024magsqlmultiagentgenerativeapproach}. 
As illustrated in Figure~\ref{fig:FIG1_toy}, however, such decompositions remain detached from the true database content—each reasoning step proceeds without feedback, resulting in schema mismatches and execution failures. 

Both directions reveal a deeper limitation: existing approaches treat reasoning as a one-way process. 
The model’s \emph{context state} (what it sees) and \emph{generation state} (how it reasons) evolve independently, preventing effective correction or adaptation. 
Bridging this divide requires a framework where these two states interact dynamically through feedback—motivating the design of our dual-state reasoning paradigm.

Building on this insight, we introduce \textbf{DSR-SQL}, a \emph{dual-state reasoning} framework that reconceptualizes Text-to-SQL as the interaction between an \emph{adaptive context state} (what the model can see) and a \emph{progressive generation state} (how it reasons). 
Rather than aligning the full schema in a single pass, DSR-SQL incrementally refines both states through execution signals: it condenses and selects schema fragments to maintain a compact, task-relevant environment, and it evolves partial SQL hypotheses based on their results. 
Complementing the failure case already shown in Figure~\ref{fig:FIG1_toy} (panel~c), Figure~\ref{fig:FIG1_toy} (panel~b) illustrates how DSR-SQL uses execution feedback to revise intermediate states, actively probe the database when mismatches are detected, and converge to the correct query. 
This dual-state formulation turns Text-to-SQL from static mapping into grounded, feedback-conditioned reasoning.
In contrast to the failure in Figure~\ref{fig:FIG1_toy} (panel~c), panel~(b) shows DSR-SQL revising states via execution feedback and converging to the correct query.
Our main contributions can be summarized as follows:

\begin{itemize}[leftmargin=*, labelsep=0.6em]
    \item We propose \textbf{DSR-SQL}, a novel \emph{dual-state reasoning} framework that reformulates Text-to-SQL as an interactive process between adaptive context management and feedback-guided generation. This formulation enables LLMs to dynamically couple perception and reasoning rather than relying on a static schema view.
    
    \item DSR-SQL introduces two complementary reasoning loops: a context-state loop that incrementally refines database visibility through schema condensation and adaptive selection, and a generation-state loop that evolves partial SQLs under execution feedback. Their co-evolution allows robust query–schema alignment even under enterprise-scale complexity.
    
    \item In a fully \emph{zero-shot} setting—without Post-training or in-context exemplars—DSR-SQL achieves strong execution accuracy on Spider~2.0-Snow and BIRD, demonstrating that test-time dual-state reasoning can match or surpass methods requiring additional supervision.
\end{itemize}

\section{Related Work}
\label{sec:related_work}
The Text-to-SQL task, aiming to convert natural language into executable SQL statements, has evolved from early sequence-to-sequence models \cite{cai2018encoderdecoderframeworktranslatingnatural,qi2022rasatintegratingrelationalstructures} and pre-trained architectures \cite{wang2021ratsqlrelationawareschemaencoding,cai2021sadga} to the current paradigm dominated by LLMs \cite{pourreza2024chasesqlmultipathreasoningpreference}. Despite these advantages, LLMs, particularly in zero-shot scenarios, often struggle with complex queries due to schema misinterpretation and logical reasoning errors \cite{li2023llm}. To mitigate this, recent zero-shot methods have moved beyond the limitations of static Chain-of-Thought prompting, instead leveraging execution feedback for post-hoc correction and validation of generated SQL candidates \cite{pourreza2023dinsqldecomposedincontextlearning,xie2024magsqlmultiagentgenerativeapproach,lee2024mcssqlleveragingmultipleprompts}. However, these methods share a fundamental limitation: they primarily treat feedback as an external signal for post-hoc optimization or validation of a complete SQL candidate \cite{qu2025share}. In contrast, DSR-SQL deeply integrates feedback into the iterative generation process itself, using it not just to correct errors, but to actively explore and understand the database's structure, thereby achieving a more fundamental alignment between user intent and database semantics.

\begin{figure*}[ht]
\centering
\includegraphics[width=1\linewidth]{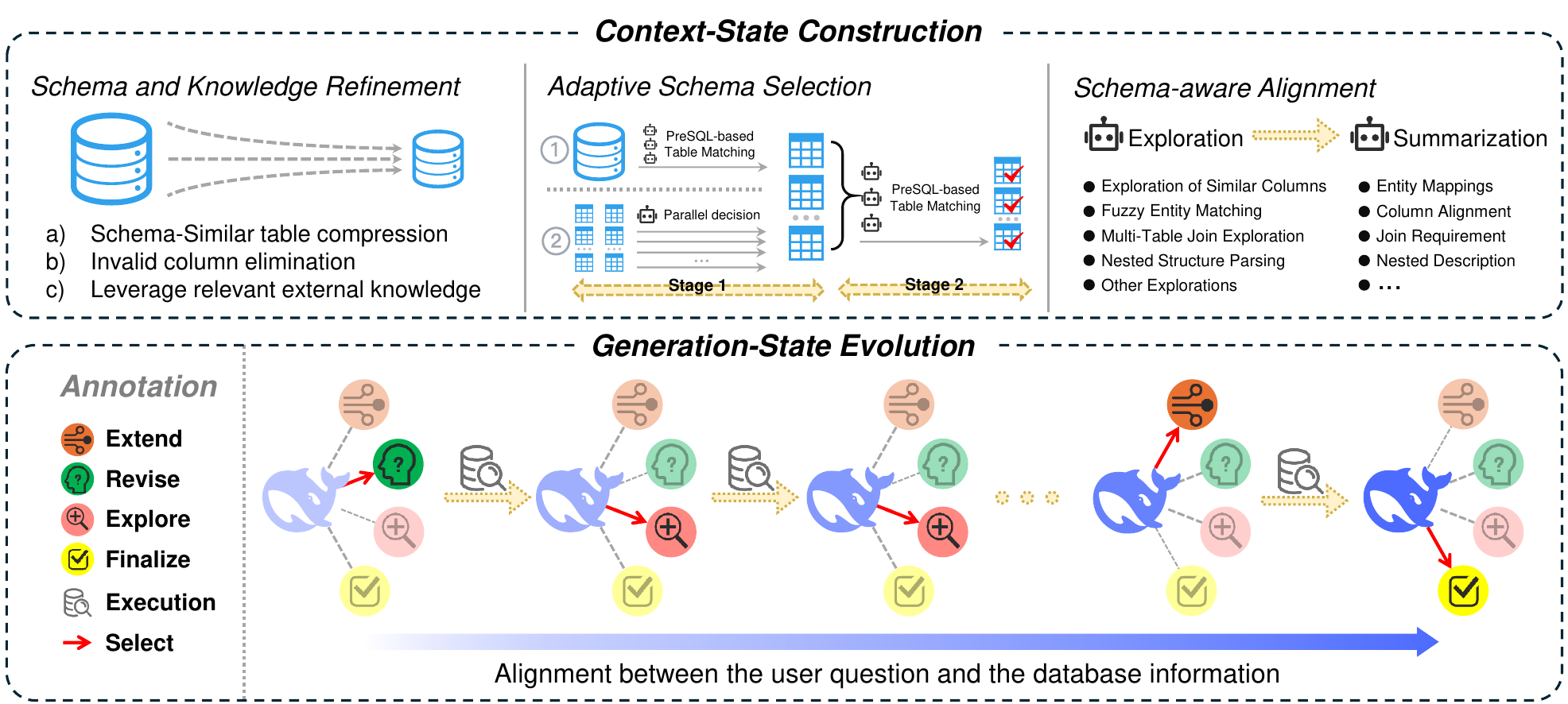}
\caption{
Overview of the DSR-SQL framework for enterprise-level Text-to-SQL. 
DSR-SQL consists of four components: 
Schema and Knowledge Refinement to handle large-scale inputs, 
Adaptive Schema Selection to identify relevant tables, 
Schema-aware Alignment to bridge linguistic and database semantics, 
and Generation-State Evolution guided by execution feedback to iteratively align user intent with query logic.
}
\label{fig2:framework}
\end{figure*}

\section{Method}

In this section, we introduce \textbf{DSR-SQL}, a structured framework designed to enhance the Text-to-SQL capabilities of general-purpose LLMs on complex enterprise-level databases. 
DSR-SQL reformulates the task as a two-state process that alternates between \textit{context construction} and \textit{generation evolution}, enabling the model to reason effectively under limited context windows. 
We begin with a formal problem formulation and then detail the function of each component within the DSR-SQL pipeline, as illustrated in Figure~\ref{fig2:framework}.

\subsection{Problem Formulation}

Given a natural-language question $\mathcal{Q}$, a database schema 
$\mathcal{S} = (\mathcal{T}, \mathcal{C})$, and optional external knowledge 
$\mathcal{K}$, the goal of Text-to-SQL is to generate a SQL query 
$\mathcal{Y}$ whose execution result is equivalent to that of the reference query on the same database.

To handle large-scale databases under context-length constraints, 
DSR-SQL divides the process into two interacting components.

\paragraph{(1) Context-State Construction.}
Before generation, the model operates within a manageable 
\emph{context state} 
$\mathcal{E} = (\mathcal{S}', \mathcal{K}')$ 
that defines the visible schema and knowledge.  
DSR-SQL constructs this state by refining lengthy inputs, selecting question-relevant schema elements, 
and aligning linguistic mentions with database semantics.  
This compact environment retains the essential information needed for accurate generation.

\paragraph{(2) Generation-State Evolution.}
Given $\mathcal{E}$, the model generates SQL through an iterative 
reasoning process that maintains a \emph{generation state} 
$s_t = (\mathcal{Y}_t, \mathcal{R}_t)$, 
where $\mathcal{Y}_t$ is a partial query and $\mathcal{R}_t$ is its execution feedback.  
The state is updated step by step until a valid and executable query is obtained, 
progressively aligning the predicted SQL with user intent and database semantics.

\subsection{Context-State Construction}

\subsubsection{Schema and Knowledge Refinement}
\label{subsec:refine}

As the first stage of \textbf{Context-State Construction}, 
DSR-SQL refines large-scale database schemas and verbose external knowledge 
to build a compact yet semantically faithful environment state 
$(\mathcal{S}', \mathcal{K}')$.  
This step ensures that subsequent reasoning operates within the context 
capacity of LLMs while preserving essential information.

\paragraph{Schema Refinement.}
We reorganize the schema by identifying and consolidating 
\textbf{structurally equivalent tables}—those that share nearly identical 
column layouts but differ only in versioning or temporal suffixes 
(e.g., \texttt{GA\_SESSIONS\_20160801}–\texttt{GA\_SESSIONS\_20170801}).  
Each series is represented by a single canonical table, accompanied by a concise 
annotation automatically generated by an LLM to describe the merged pattern 
(see Appendix~\ref{app:table_des}).  
Columns that contain consistently null or uninformative values are pruned, 
yielding a reduced schema $\mathcal{S}' = (\mathcal{T}', \mathcal{C}')$ with 
$|\mathcal{T}'| \ll |\mathcal{T}|$.

\paragraph{Knowledge Refinement.}
Similarly, verbose external knowledge $\mathcal{K}$ is summarized into 
$\mathcal{K}'$ through LLM-based extraction and synthesis, 
retaining only information directly relevant to the question—such as 
units, enumerations, or temporal ranges.

\begin{figure}[ht]
\centering
\includegraphics[width=1\linewidth]{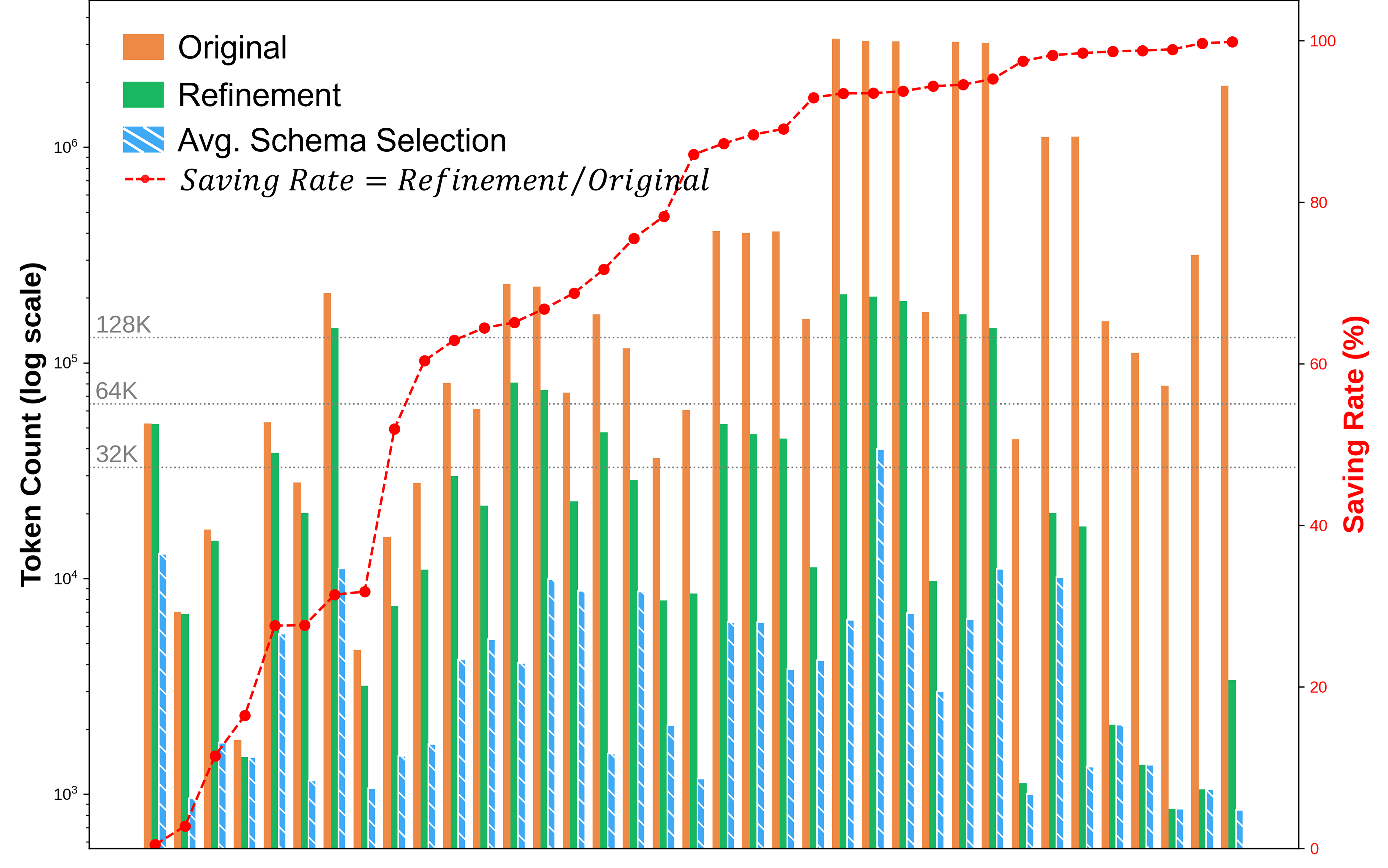}
\caption{
Schema token counts (M-Schema format) for selected databases in Spider 2.0-Snow. The comparison shows the original schema size (orange), the size after \textit{Schema and Knowledge Refinement} (green), and the average per-query size after \textit{Adaptive Schema Selection} (blue-and-white striped).
}
\label{fig3:IC}
\end{figure}

As shown in Figure~\ref{fig3:IC}, most refined M-schemas after this reduction 
fall below 128K tokens, enabling efficient context utilization for 
later stages of Context-State Construction.

\subsubsection{Adaptive Schema Selection}
\label{subsec:selection}

As the second stage of \textbf{Context-State Construction}, 
DSR-SQL dynamically selects a subset of the refined schema 
that is most relevant to the input question $\mathcal{Q}$.  
Given the compressed schema $\mathcal{S}'$, the model identifies 
a task-specific subset $\mathcal{S}_{\text{sub}} \subseteq \mathcal{S}'$ 
through an adaptive exploration strategy that balances 
coverage and efficiency.

\paragraph{Global Exploration.}
When the schema size remains within the context limit, 
the model performs global exploration: 
an LLM infers potential SQL structures over the full schema 
and collects tables that appear across high-confidence candidates 
as the relevant subset $\mathcal{S}_{\text{sub}}$.

\paragraph{Partitioned Exploration.}
For larger databases that exceed the context window, 
DSR-SQL adopts a partitioned exploration strategy to ensure 
robust table selection under varying schema scales. 
It evaluates each table independently through 
query--table similarity prompts and retains only 
those predicted to be relevant.

Through this adaptive selection process, most per-query database schemas are thus reduced 
to under 32K tokens (see Figure~\ref{fig3:IC}), greatly alleviating the context burden.  
The detailed procedure is provided in 
Appendix~\ref{app:Dynamic_tab} as pseudocode.

\subsubsection{Schema-aware Alignment}
\label{subsec:alignment}

The final stage of \textbf{Context-State Construction} establishes 
an initial semantic grounding between the natural-language question 
and the selected schema $\mathcal{S}_{\text{sub}}$.  
To obtain a content-aware alignment prior, 
DSR-SQL performs a two-phase procedure of lightweight exploration 
and summarization.

\paragraph{Exploration.}
An LLM first generates a small set of simple \emph{probing queries} 
(e.g., \texttt{SELECT COUNT(*)}, \texttt{SELECT DISTINCT}) 
to inspect key value distributions and entity relationships within 
$\mathcal{S}_{\text{sub}}$.  
The execution results of these queries reveal 
frequent values, typical ranges, and common join patterns.

\paragraph{Summarization.}
The obtained results are then synthesized by another LLM prompt 
into a concise description $\mathcal{K}_{\text{align}}$ that captures 
salient domain information (e.g., enumerations, units, synonyms).  
This alignment prior provides the generator with explicit semantic cues, 
helping it avoid redundant or inefficient reasoning steps 
in the subsequent generation stage 
(see Appendix~\ref{app:sqlpath} for an illustration).

\subsection{Generation-State Evolution}
\label{subsec:generation}

After constructing the contextual environment, 
DSR-SQL performs \textbf{generation-state evolution}, 
an iterative reasoning process guided by execution feedback.  
Rather than generating the entire SQL in a single pass, 
the model maintains a generation state 
$s_t = (\mathcal{Y}_t, \mathcal{R}_t)$, 
where $\mathcal{Y}_t$ is a partial query and $\mathcal{R}_t$ is its execution result.  
At each step, this state evolves based on feedback signals and the prepared context 
$\mathcal{X}' = (\mathcal{Q}, \mathcal{S}_{\text{sub}}, \mathcal{K}', \mathcal{K}_{\text{align}})$.  

This process can be conceptualized as a state-driven reasoning procedure governed by four recurrent actions, each triggered by distinct feedback conditions:

\paragraph{Extend.}
When the current execution $\mathcal{R}_t$ yields valid and informative results, 
the model extends the partial query to handle the next aspect of the user question, 
progressively composing a more complete SQL structure.

\paragraph{Revise.}
If execution feedback reveals a semantic or logical inconsistency, 
the model revises the existing partial query, correcting the reasoning path 
based on error signals from the database.

\paragraph{Explore.}
When execution produces unexpected or empty outputs, 
the model issues short exploratory queries to probe relevant value ranges 
or relational connections before resuming the main reasoning path.

\paragraph{Finalize.}
Once all sub-goals have been resolved and execution outcomes align with 
the user’s intent, the model synthesizes a coherent and executable SQL query, 
terminating the evolution process.

Through these feedback-driven transitions, 
DSR-SQL progressively aligns natural-language intent with database semantics, 
forming a directed, acyclic reasoning trajectory that converges to the final SQL query.  
The detailed prompting design and pseudocode are provided 
in Appendix~\ref{app:FSM_SQL_Generation}.

\section{Experiments}

\subsection{Benchmarks}
To comprehensively evaluate the performance of our proposed DSR-SQL framework, we conducted extensive experiments on two widely-used, cross-domain Text-to-SQL benchmarks: Spider 2.0\cite{lei2024spider2} and BIRD\cite{li2023llm}. Spider 2.0 is an emerging and highly challenging benchmark designed to simulate real-world, enterprise-level application scenarios. Its database schemas are exceptionally complex, averaging approximately 800 columns per database, and the required SQL queries present a formidable challenge in terms of complexity. For this evaluation, we focus on its sub-task, Spider 2.0-Snow, which requires generating SQL in the Snowflake dialect. In contrast, the BIRD benchmark places a greater emphasis on understanding fine-grained schema details, such as distinguishing between similarly named columns and handling minor typographical errors in questions, making it equally challenging.

\subsection{Evaluation Metrics}
Following the standard in the field, we adopt the widely recognized Execution Accuracy (EX) as our core performance metric. It is important to note that the definition of EX actually differs between the two benchmarks. BIRD employs a strict evaluation standard, requiring the execution results of a generated query to be an exact match with the gold standard in both content and column order. Spider 2.0-Snow, however, utilizes a more lenient evaluation script. Acknowledging the potential ambiguity in user intent, its evaluation only requires that the returned results contain all the core information of the gold answer, without constraining the order of columns or the presence of any additional ones, thereby minimizing false negatives in the evaluation process.

\subsection{Implementation Details}
Our experiments employ several LLMs from the DeepSeek and Qwen series, including DeepSeek-R1-0528 (DS-R1), DeepSeek-V3-0324 (DS-V3), DeepSeek-V3.1 (DS-V3.1), and Qwen3-30B-A3B-Instruct-2507 (Qwen3-30B). Unless otherwise specified, the decoding temperature was set to 0.2. For the Generation-State Evolution, the maximum number of iterations is set to 20 for Snow and 10 for BIRD development datasets, reflecting the differing query complexities of the datasets. In addition, a query correction loop (up to 5 iterations) following \citet{pourreza2024chasesqlmultipathreasoningpreference} was applied to all methods to fix minor syntactic errors. For more details on the comparison with baselines, see Appendix ~\ref{app:baseline}.

\subsection{Comparison Results}

\begin{table}[htbp]
    \centering
    \begin{tabular}{lcc} 
        \toprule
        \textbf{Method} & \textbf{Voting} & \textbf{EX (\%)} \\
        \midrule
        \makecell[l]{Spider-Agent + GPT-4o} & \textcolor{red}{\ding{55}} & 10.05 \\
        \addlinespace
        \makecell[l]{Spider-Agent + DS-R1} & \textcolor{red}{\ding{55}} & 10.79 \\
        \addlinespace
        \makecell[l]{ReFORCE + GPT-4o} & \textcolor{green!60!black}{\ding{51}} & 20.84 \\
        \addlinespace
        \makecell[l]{ReFORCE + DS-R1} & \textcolor{green!60!black}{\ding{51}} & 29.25 \\
        \midrule
        \makecell[l]{DSR-SQL + DS-V3} & \textcolor{red}{\ding{55}} & 22.67 \\
        \addlinespace
        \makecell[l]{DSR-SQL + DS-R1} & \textcolor{red}{\ding{55}} & \textbf{35.28} \\
        \bottomrule
    \end{tabular}
    
    \caption{Comparison on Spider 2.0-Snow. The "Voting" column indicates whether multi-path voting was used: a \textcolor{green!60!black}{\ding{51}} denotes that voting was applied, while a \textcolor{red}{\ding{55}} denotes that it was not.}
    \label{tab:spider_snow}

\end{table}

\begin{table*}[ht]
    \centering
    \renewcommand{\arraystretch}{0.9} 
    \begin{tabular}{l S[table-format=2.2]
                      S[table-format=2.2]
                      S[table-format=2.2]
                      S[table-format=2.2]}
        \toprule
        \multicolumn{1}{c}{\multirow[c]{2}{*}{Method}} & \multicolumn{4}{c}{EX (\%)} \\
        \cmidrule(lr){2-5}
        & {Simple} & {Moderate} & {Challenging} & {Total} \\
        \midrule
        \multicolumn{5}{l}{\textit{Hybrid}} \\
        CHASE-SQL + Gemini     & {--}           & {--}           & {--}           & 74.90 \\
        XiYan-SQL + Multi-model  & {--}           & {--}           & {--}           & 73.34 \\
        AskData + GPT4o         & {--}           & {--}           & {--}           & 75.36 \\
        Contextual-SQL         & {--}           & {--}           & {--}           & 73.50 \\
        \midrule
        \multicolumn{5}{l}{\textit{Post-training}} \\
        XiYanSQL-QwenCoder-32B  & 73.62          & 60.34          & 51.72          & 67.54 \\
        Arctic-Text2SQL-R1-7B   & 73.62          & 62.50          & 57.24          & 68.71 \\
        Databricks-RLVR-32B     & {--}           & {--}           & {--}           & 70.80 \\
        OmniSQL-32B             & 71.78          & 54.31          & 50.34          & 64.47 \\
        \midrule
        \multicolumn{5}{l}{\textit{In-context Learning}} \\
        DAIL-SQL + GPT4         & 62.49          & 43.44          & 38.19          & 54.43 \\
        MCS-SQL + GPT4          & 70.40          & 53.10          & 51.40          & 63.40 \\
        RSL-SQL + DS-V3.1       & 72.11          & 59.27          & 53.79          & 66.49 \\
        OpenSearch-SQL + DS-V3.1& 74.05          & 62.07          & 57.24          & 68.84 \\
        \midrule
        \multicolumn{5}{l}{\textit{Zero-shot}} \\
        DIN-SQL + GPT4          & {--}           & {--}           & {--}           & 50.72 \\
        MAC-SQL + GPT4          & {--}           & {--}           & {--}           & 55.54 \\
        MAG-SQL + DS-V3.1       & 71.24          & 59.27          & 55.17          & 66.10 \\
        DS-V3.1                 & 64.54          & 50.43          & 49.66          & 58.87 \\
        Qwen3-30B               & 64.00          & 48.06          & 38.62          & 56.78 \\
        \midrule
        DSR-SQL + Qwen3-30B     & 70.70          & 59.27          & 53.10          & 65.58 \\
        DSR-SQL + DS-V3.1       & \textbf{72.65} & \textbf{61.21} & \textbf{63.45} & \textbf{68.32} \\
        \bottomrule
    \end{tabular}
    \caption{Comparison of Methods on the BIRD Development dataset.}
    \label{tab:bird_results}
\end{table*}

\paragraph{Results on Spider2-Snow} Table~\ref{tab:spider_snow} shows the performance comparison on the Snow benchmark. With a single generation path, DSR-SQL paired with DS-R1 achieves an EX of 35.28\%. This result represents a 6.03\% absolute improvement over ReFORCE, which relies on an 8-path voting mechanism. Notably, our framework with the DS-V3 model also surpasses ReFORCE paired with the much stronger GPT-4o model (20.84\%). These findings demonstrate the effectiveness of the DSR-SQL framework in handling complex databases without requiring multi-path generation. We conduct a detailed error analysis on the Snow benchmark; see Appendix~\ref{app:error_ana} for details.

\begin{table}[ht]
    \centering
    \begin{tabular}{lc} 
        \toprule
        \textbf{Method} & \textbf{EX (\%)} \\
        \midrule
        DSR-SQL (Full)                & 35.28 \\
        \midrule 
        DSR-SQL w/o SKR            & 22.48 {\small ($\downarrow$ 12.80)} \\
        DSR-SQL w/o ASS    & 24.49 {\small ($\downarrow$ 10.79)} \\
        DSR-SQL w/o SAA             & 31.80 {\small ($\downarrow$ 3.48)} \\
        DSR-SQL w/o GSE             & 27.79 {\small ($\downarrow$ 7.49)} \\
        \bottomrule
    \end{tabular}
    \caption{Ablation study of different components in DSR-SQL on the Spider 2.0-Snow dataset.}
    \label{tab:ablation-study}
\end{table}

\paragraph{Results on BIRD} In \textbf{Table~\ref{tab:bird_results}}, we provide a detailed performance comparison of DSR-SQL against mainstream Text-to-SQL frameworks on the BIRD development set. To ensure a fair assessment of each method's foundational performance, the number of samples was uniformly set to one for all baselines.

The results demonstrate that, under a zero-shot setting, DSR-SQL achieves strong execution accuracies of 68.32\% and 65.58\% when applied to two leading LLMs (DS-V3.1 and Qwen3-30B). This performance surpasses all existing zero-shot methods and is comparable to—or even exceeds—most approaches based on Post-training and in-context examples. While a gap remains compared to certain hybrid methods, it is important to note that those approaches often depend on closed-source models and proprietary Post-training datasets, which substantially increase computational costs and raise potential data privacy concerns.

\subsection{Ablation and Module Analysis}
\subsubsection{Overall Ablation Study}
We performed an ablation study on the Spider 2.0-Snow benchmark based on the DS-R1 model (see Table~\ref{tab:ablation-study}) to measure each module's contribution to DSR-SQL. The results show that every module is essential. The Schema and Knowledge Refinement (SKR) module is foundational, tackling the core challenge of limited context capacity by condensing oversized schemas. Building on this, the Adaptive Schema Selection (ASS) and Schema-aware Alignment (SAA) components are crucial for constructing a compact, semantically faithful environment by selecting relevant structures and grounding linguistic terms in database semantics. Omission of any module produces measurable performance drops of varying magnitudes. Notably, the Generation-State Evolution (GSE) module leverages this carefully constructed context state, using feedback-guided state transitions to incrementally align user intent with the database. This approach attains a higher performance ceiling than conventional divide-and-conquer methods, corroborating the effectiveness of our dual-state design.

\subsubsection{Analysis of Adaptive Schema Selection}
We evaluate the efficiency of our Adaptive Schema Selection approach across multiple dimensions, with results presented \textbf{in Table~\ref{tab:Table_Matching}.} On the challenging Snow dataset, our ASS module achieves a recall of 91.13\%, a significant improvement over the 62.59\% achieved by the matching method in ReFoRCE. This higher accuracy comes at the cost of increased token consumption (34.35K vs. 23.23K). However, our approach reduces the average number of LLM calls from 13.29 to 6.06, indicating a more efficient reasoning process. On the BIRD dataset, ASS achieves over 98\% recall with both tested models, demonstrating its robustness.

\newlength{\datasetcolwidth}
\settowidth{\datasetcolwidth}{\textbf{Dataset}}

\begin{table*}[h!]
\centering 
\begin{threeparttable}

    \begin{tabular}{ccccccc}
    \toprule
    \textbf{Dataset} & \textbf{Method} & \textbf{Model} & \textbf{Precision (\%)} & \textbf{Recall (\%)} & \textbf{Avg. Tokens} & \textbf{Avg. LLM Calls} \\ 
    \midrule
    \multirow{2}{*}{\parbox{\datasetcolwidth}{\centering\textbf{Snow}}} & ReFORCE & DS-V3    & 58.09             & 62.59          & \textbf{23.23K}                  & 13.29                   \\
                                           & DSR-SQL     & DS-V3    & \textbf{75.62}    & \textbf{91.13} & 34.35K                  & \textbf{6.06}           \\ 
    \midrule
    \multirow{2}{*}{\parbox{\datasetcolwidth}{\centering BIRD}}         & DSR-SQL     & DS-V3.1  & 63.07             & 98.41          & \textbf{3.67K}                   & \textbf{1.00}           \\
                                           & DSR-SQL     & Qwen3-30B & \textbf{64.59}    & \textbf{98.83} & 3.80K                   & \textbf{1.00}           \\ 
    \bottomrule
    \end{tabular}
\end{threeparttable}
\caption{Performance Analysis of Adaptive Schema Selection. We set the model temperature to 1.2 and generate three samples to improve recall. ReFoRCE defaults to using the full schema, invoking its table matching method only when the context exceeds 200K tokens. Thus, its matching cost is low but has minimal impact on its final SQL accuracy. Token counts may vary slightly across models.}
\label{tab:Table_Matching}
\end{table*}

\subsubsection{Analysis of Schema-aware Alignment}
We conducted an ablation study on the Schema-aware Alignment (SAA) module to assess its contribution to the overall framework. To ensure a fair and consistent comparison, we reused the intermediate results from the Exploration phase across all relevant experimental setups. The results are presented in Table~\ref{tab:Pre_ALIGNMENT_ANA}.

The analysis reveals several key findings. First, removing the SAA module entirely results in a substantial performance drop of approximately 3.5\%. Second, providing the model with only the raw \texttt{(SQL, execution result)} pairs from the Exploration phase yields performance considerably lower than that of the full framework. This suggests that without the Summarization step, the LLM struggles to extract deep insights from the long-context, unprocessed data. Third, using Summarization alone, without the data-grounded context from Exploration, leads to a significant performance drop and can even be detrimental, yielding results comparable to a simple divide-and-conquer approach (see Appendix~\ref{app:DC}). This is likely due to model hallucinations caused by the lack of concrete, exploratory information.

Notably, even a baseline divide-and-conquer method improves when augmented with our Schema-aware Alignment module, further underscoring the positive guiding role the module plays in SQL generation.

\begin{table}[h!]
\centering 
\begin{threeparttable}

    \begin{tabular}{lc}
    \toprule
    \textbf{Method} & \textbf{EX (\%)} \\ 
    \midrule
    DSR-SQL (Full) & 35.28 \\ 
    DSR-SQL w/o SAA & 31.8 {\small ($\downarrow$ 3.48)} \\
    DSR-SQL w/o Summarization & 32.35 {\small ($\downarrow$ 2.93)} \\
    DSR-SQL w/o Exploration & 28.70 {\small ($\downarrow$ 6.58)} \\
    divide-and-conquer w SAA & 27.79 {\small ($\downarrow$ 7.49)} \\
    divide-and-conquer w/o SAA & 25.05 {\small ($\downarrow$ 10.23)} \\
    \bottomrule
    \end{tabular}
\end{threeparttable}
\caption{Evaluating the performance impact of the Schema-aware Alignment module on SQL generation on the Spider2.0-Snow dataset.}
\label{tab:Pre_ALIGNMENT_ANA}
\end{table}

\subsubsection{Analysis of Generation-State Evolution}
The Generation-State Evolution module plays a pivotal role in DSR-SQL, enabling dynamic alignment with and understanding of complex databases to facilitate accurate SQL generation. We conducted a corresponding ablation study on both the Spider 2.0-Snow and BIRD development datasets to quantify its impact.

The results demonstrate that, as shown in Table~\ref{tab:SQL_Generation_comparison}, compared to a static divide-and-conquer approach, our dynamic method holds a clear advantage on complex, large-scale databases like Spider 2-Snow, with a performance improvement of approximately 7.49\%. On the BIRD development dataset, our method also exhibits superior performance in resolving ambiguous and challenging user queries, highlighting the generalizability of our approach.

\begin{table}[ht]
\centering 
\begin{threeparttable}

    \setlength{\tabcolsep}{3pt} %
    \begin{tabular}{cccc}
    \toprule
    \textbf{Dataset} & \textbf{Method} & \textbf{Model} & \textbf{EX (\%)} \\ 
    \midrule
    \multirow{2}{*}{\parbox{\datasetcolwidth}{\centering Snow}} & divide-and-conquer     & \multirow{2}{*}{DS-R1}    & 27.79             \\
                                           & DSR-SQL                &                        & \textbf{35.28}    \\ 
    \midrule
    \multirow{2}{*}{\parbox{\datasetcolwidth}{\centering BIRD}}         & divide-and-conquer     & \multirow{2}{*}{DS-V3.1}  & 63.17             \\
                                           & DSR-SQL                &                        & \textbf{68.32}    \\ 
    \bottomrule
    \end{tabular}
\end{threeparttable}
\caption{Performance Analysis of the Generation-State Evolution module, where divide-and-conquer utilizes the Schema-aware Alignment module.}
\label{tab:SQL_Generation_comparison}
\end{table}

Furthermore, we analyzed the accuracy rates across different generation paths within the SQL Generation module. We categorize the generation process into three distinct path types (note: this classification is independent of the dataset's official difficulty ratings):
\begin{itemize}
    \item \textbf{Straightforward Path:} Paths that construct the final SQL using only \textbf{Extend} and \textbf{Finalize} actions. This indicates that the LLM has high confidence in its solution and requires no additional information.
    \item \textbf{Refinement Path:} Paths composed of a combination of \textbf{Extend}, \textbf{Revise}, and \textbf{Finalize} actions. This signifies that intermediate execution results may have partially deviated from the user's requirements.
    \item \textbf{Exploratory Path:} Paths that utilize the \textbf{Explore} action. This suggests that the user's question is particularly difficult or ambiguous, necessitating in-depth database exploration to formulate the correct SQL.
\end{itemize}

\begin{figure}
\centering
\includegraphics[width=1\linewidth]{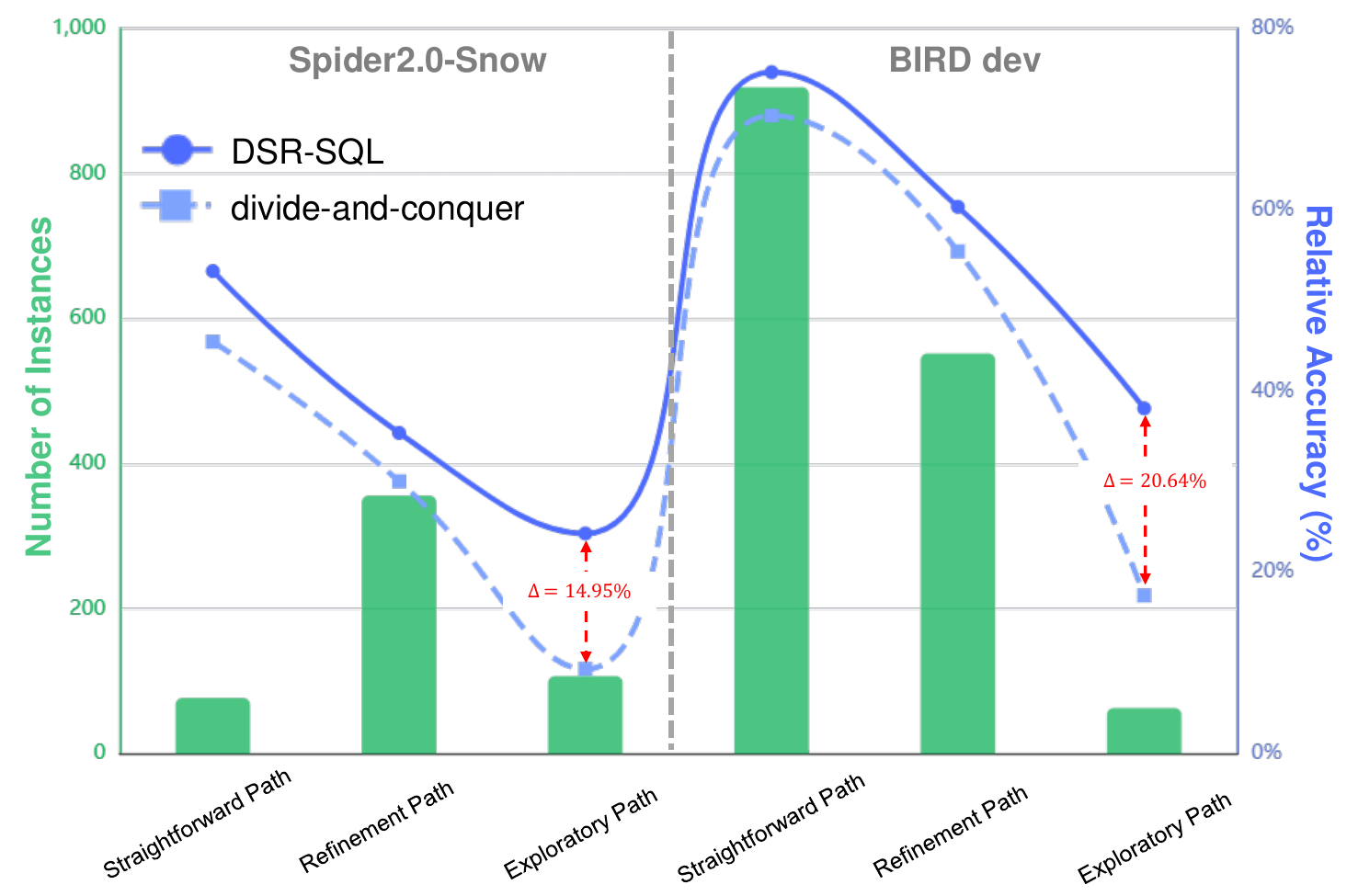}
\caption{Accuracy comparison of the SQL Generation module by generation path type on the Spider 2.0-Snow and BIRD dev datasets, with divide-and-conquer making use of the Schema-aware Alignment module.}
\label{fig:path_ana}
\end{figure}

The detailed results are shown in Figure~\ref{fig:path_ana}. DSR-SQL consistently outperforms across all path types, with the largest gains on the \textbf{Exploratory Path}, improving relative accuracy by 14.95\% and 20.64\% on Spider 2-Snow and BIRD, respectively.

\section{Conclusions}
In this work, we introduced DSR-SQL, a dual-state reasoning framework that models Text-to-SQL as the interaction between an adaptive context state and a progressive generation state. This formulation enables LLMs to maintain coherent reasoning over complex enterprise databases, addressing the challenges of context limitation and schema grounding.
Experiments on Spider 2.0-Snow and BIRD demonstrate that DSR-SQL achieves strong zero-shot performance without Post-training or in-context examples.
Beyond these results, our findings suggest that balancing how models manage external context and evolve internal reasoning states can be an effective path toward more reliable structured reasoning.
We hope this perspective encourages further exploration into state-aware reasoning frameworks across other data-intensive tasks.

\section*{Limitations}
While our approach demonstrates superior performance over current SOTA methods on Spider 2.0-Snow and BIRD in a zero-shot scenario, it still lags behind certain techniques that leverage Post-training, in-context examples, or multi-path generation. Addressing this performance disparity will be a key focus of future work.

\bibliography{custom}

\appendix
\section{Table description generation}
\label{app:table_des}
For the tables that underwent Schema Refinement, we used an LLM to generate a simple table description. Specifically, we fed all table names and their corresponding column descriptions for a set of structurally equivalent tables into a prompt, instructing the LLM to generate a concise description that summarizes them. An example is provided in Figure \ref{fig:Example_of_table_des}.
\begin{figure}[htbp]
\centering
\begin{promptbox}[Database Name: GA360]
The naming convention for these tables with the same structure is: 
\begin{enumerate}[leftmargin=*] 
    \item \textit{GA\_SESSIONS\_[DATE]}, where [DATE] represents the date of the session data in `YYYYMMDD` format, \textbf{ranging from 20160801 to 20170801}, covering a full 13-month period from August 1, 2016, to August 1, 2017.
    
    \item These tables represent daily partitions of Google Analytics session-level data, capturing metrics such as pageviews, sessions, bounces, hits, and session duration for each day.
    
    \item All tables share the same core schema, but an important schema change occurs starting July 1, 2017: \textbf{tables from 20170701 through 20170801 include an additional column \texttt{clientId}, which is absent in earlier tables.} This suggests an enhancement to track client-level identifiers for user journey analysis starting in July 2017.
\end{enumerate}

\end{promptbox}
\caption{Example of Table description}
\label{fig:Example_of_table_des}
\end{figure}
\section{M-Schema}
\label{app:mschema}
M-Schema, as proposed by \citet{liu2025xiyansqlnovelmultigeneratorframework}, is a semi-structured representation of a database schema. Compared to simple database creation statements, it encapsulates richer information, such as value examples and column descriptions. In this work, our configuration largely follows the original implementation, with two key modifications. First, for the value examples of each column, we first deduplicate them and then randomly select up to three distinct values, with each example truncated to a maximum length of 50 characters. Second, for tables that have been compressed (as described in our preprocessing step), we introduce a new \textit{\# Table Description} field to articulate their specific characteristics. An illustrative example of our modified M-Schema format is shown in Figure~\ref{fig:Example_of_M-Schema}.
\begin{figure*}[htbp]
\centering

\begin{promptbox}[M-Schema: GA360]
[DB\_ID] GA360

\# Table: GA360.GOOGLE\_ANALYTICS\_SAMPLE.GA\_SESSIONS\_20160801

[

(visitNumber: NUMBER, The session number for this user. If this is the first session, then this is set to 1., Examples: [1, 3, 30]),

(visitId: NUMBER, An identifier for this session. This is part of the value usually stored as the \_utmb cookie. This is only unique to the user. For a completely unique ID, you should use a combination of \texttt{\`{}}fullVisitorId\texttt{\`{}} and \texttt{\`{}}visitId\texttt{\`{}}., Examples: [1470046245, 1470072494, 1470078988]),

(visitStartTime: NUMBER, The timestamp (expressed as POSIX time)., Examples: [1470046245, 1470072494, 1470078988]),
(date: TEXT, The date of the session in YYYYMMDD format., Examples: [20160801]),

…

] 

\# Table Description: The naming convention for these tables with the same structure is:
GOOGLE\_ANALYTICS\_SAMPLE.GA\_SESSIONS\_[DATE], where [DATE] represents the date of the session data in \texttt{\`{}}YYYYMMDD\texttt{\`{}} format, ranging from 20160801 to 20170801, covering a full 13-month period from August 1, 2016, to August 1, 2017. 

…
\end{promptbox}
\caption{Example of M-Schema}
\label{fig:Example_of_M-Schema}
\end{figure*}

\section{Example of Different Generation-State Evolution Paths}
\label{app:sqlpath}
\begin{figure*}[htbp]
\centering
\includegraphics[width=1\linewidth]{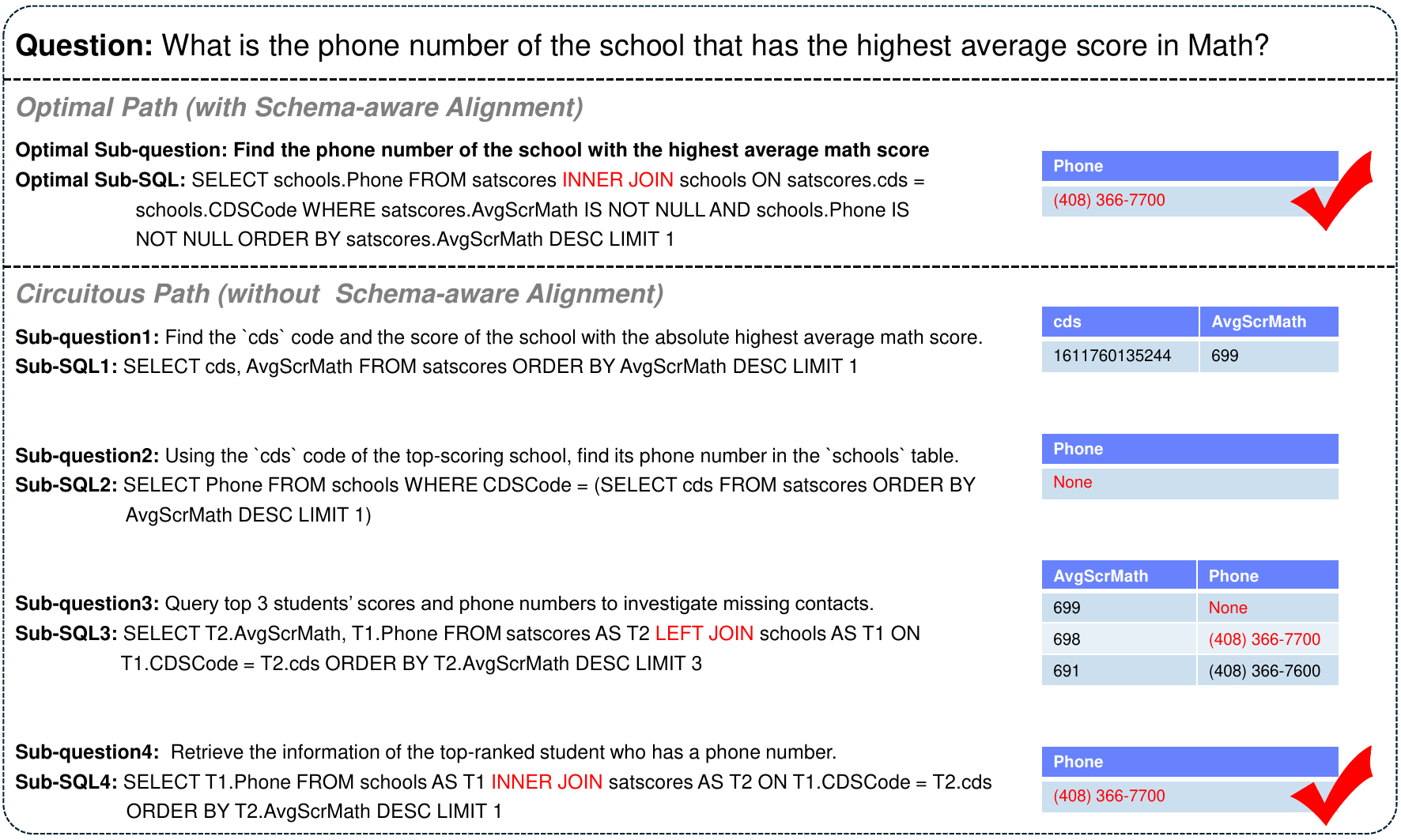}
\caption{Example of Decomposition Paths: With vs. Without the Schema-aware Alignment module}
\label{fig:Decomposition_Paths}
\end{figure*}

This section presents an example of a single Text-to-SQL task being solved via two distinct evolution paths, as shown in Figure~\ref{fig:Decomposition_Paths}. Although both paths ultimately yield the correct result, the ``Circuitous Path'' requires multiple steps to arrive at the solution. In contrast, the ``Optimal Path'' generates the answer in a single step by leveraging the implicit null-value filtering of an \texttt{INNER JOIN}. This comparison highlights the importance of our Schema-aware Alignment module in guiding the model toward more efficient and direct reasoning.

\section{divide-and-conquer Baseline}
\label{app:DC}
This section details the divide-and-conquer baseline method used for comparison in our main experiments.
For the BIRD dataset, we adopted a state-of-the-art Chain-of-Thought (CoT) prompting strategy similar to that of \citet{pourreza2024chasesqlmultipathreasoningpreference}, which includes a fixed set of few-shot examples to guide the LLM in generating the target SQL.
For the Spider 2.0-Snow dataset, our baseline model was the DeepSeek R1. In accordance with the official guidelines for this model, we did not use few-shot examples. we prompted the LLM to directly follow a \texttt{sub-question $\rightarrow$ sub-SQL $\rightarrow$ final SQL} decomposition format to generate the answer.
An example of the CoT decomposition used for a sample from the BIRD development set is provided in the Figure~\ref{fig:sql_divide_conquer_example} .

\begin{figure}[htbp]
\centering
\begin{promptbox}[divide-and-conquer Example]
\textbf{Question}: List the city and state of members enrolled under electrical and computer engineering department.  

\textbf{Evidence}: ``Electrical and Computer Engineering Department'' is the department; members enrolled refers to position = `Member`.

\bigskip
\textbf{1. divide-and-conquer:}

\textbf{Analysis:} The question asks for \texttt{city} and \texttt{state} from the \texttt{zip\_code} table, filtered to members who are enrolled (\texttt{position = 'Member'}) and whose major's department is ``Electrical and Computer Engineering Department''. This requires joining \texttt{member}, \texttt{major}, and \texttt{zip\_code} tables.

\textbf{Pseudo SQL:} 

SELECT T3.city, T3.state 
FROM member AS T1 
INNER JOIN major AS T2 ON T1.link\_to\_major = T2.major\_id 
INNER JOIN zip\_code AS T3 ON T1.zip = T3.zip\_code 
WHERE [enrolled member] AND [electrical and computer engineering department]

\bigskip

\textbf{Sub-question 1:} enrolled member

\textbf{Analysis:} Filter on \texttt{position} column in \texttt{member} table.

\textbf{Pseudo SQL:} T1.position = 'Member'

\bigskip

\textbf{Sub-question 2:} electrical and computer engineering department

\textbf{Analysis:} Filter on \texttt{department} column in \texttt{major} table.

\textbf{Pseudo SQL:} T2.department = 'Electrical and Computer Engineering Department'

\bigskip

\textbf{2. Assembling SQL:}

\textbf{Sub-question 1 (enrolled member):} T1.position = 'Member'

\bigskip

\textbf{Sub-question 2 (electrical and computer engineering department):} T2.department = 'Electrical and Computer Engineering Department'

\bigskip

\textbf{Main Question (city and state of members):}

SELECT T3.city, T3.state 
FROM member AS T1 
INNER JOIN major AS T2 ON T1.link\_to\_major = T2.major\_id 
INNER JOIN zip\_code AS T3 ON T1.zip = T3.zip\_code 
WHERE T1.position = 'Member' 
AND T2.department = 'Electrical and Computer Engineering Department'

\bigskip

\textbf{3. Simplification and Optimization:}

The query uses efficient INNER JOINs and straightforward WHERE conditions. No further optimization is needed.

\bigskip
\verb|<sql>|

SELECT T3.city, T3.state FROM member AS T1 JOIN major AS T2 ON T1.link\_to\_major = T2.major\_id JOIN zip\_code AS T3 ON T1.zip = T3.zip\_code WHERE T1.position = 'Member' AND T2.department = 'Electrical and Computer Engineering Department'

\verb|</sql>|
\end{promptbox}

\caption{An example of the divide and conquer method.}
\label{fig:sql_divide_conquer_example}
\end{figure}

\section{Adaptive Schema Selection Algorithm}
\label{app:Dynamic_tab}
In this section, we present the pseudocode of the Adaptive Schema Selection algorithm; see Algorithm~\ref{app:Dynamic_tab} for details.
\begin{algorithm}[H]
\caption{Adaptive Schema Selection}
\label{alg:Adaptive_Schema_Selection}
\begin{algorithmic}[1]

\Require Question $\mathcal{Q}$, refined schema $\mathcal{S'}$, refined knowledge $\mathcal{K'}$, LLM $\mathcal{M}$, samples $k$, threshold $\theta_{\text{max}}$
\Ensure Question-relevant schema subset $\mathcal{S}_{\text{sub}}$

\State $\mathcal{T}' \leftarrow \text{GetTab}(\mathcal{S}')$ 

\If{$|\mathcal{T}'| = 1$}
    \State $\mathcal{T}_{\text{final}} \leftarrow \mathcal{T}'$ 
\Else

    \State $\mathcal{S}_{w} \leftarrow \text{ReprSch}(\mathcal{T}', \text{'M'})$ 
    \If{$\text{TokCnt}(\mathcal{S}_{w}) > \theta_{\text{max}}$}
        \State $\mathcal{S}_{w} \leftarrow \text{ReprSch}(\mathcal{T}', \text{'DDL'})$ 
        \If{$\text{TokCnt}(\mathcal{S}_{w}) > \theta_{\text{max}}$}

            \State $\mathcal{T}_{c} \leftarrow \emptyset$
            \For{\textbf{each} table $\mathcal{T}_j \in \mathcal{T}'$ \textbf{in parallel}}
                \State $\mathcal{Y}_j \sim P_{\mathcal{M}}(\cdot|\mathcal{Q}, \mathcal{K'}, \text{SchOf}(\mathcal{T}_j))$
                \State $\mathcal{T}_{c} \leftarrow \mathcal{T}_{c} \cup \text{TabFromSQL}(\mathcal{Y}_j)$
            \EndFor
        \Else
            \State $\mathcal{T}_{c} \leftarrow \text{SampTab}(\mathcal{Q}, \mathcal{K'}, \mathcal{S}_{w}, \mathcal{M}, k)$
        \EndIf
    \Else
        \State $\mathcal{T}_{c} \leftarrow \text{SampTab}(\mathcal{Q}, \mathcal{K'}, \mathcal{S}_{w}, \mathcal{M}, k)$
    \EndIf

    \State $\mathcal{S}_{c} \leftarrow \text{ReprSch}(\mathcal{T}_{c})$
    \State $\mathcal{T}_{\text{final}} \leftarrow \text{SampTab}(\mathcal{Q}, \mathcal{K'}, \mathcal{S}_{c}, \mathcal{M}, k)$
\EndIf

\State $\mathcal{S}_{\text{sub}} \leftarrow \text{ReprSchCol}(\mathcal{T}_{\text{final}})$ 
\State \textbf{return} $\mathcal{S}_{\text{sub}}$ 

\end{algorithmic}
\end{algorithm}

\section{Generation-State Evolution}
This section provides the pseudocode of Generation-State Evolution; see Algorithm~\ref{alg:Generation_State_Evolution} for details.
\label{app:FSM_SQL_Generation}
\begin{algorithm}[H]
\caption{Generation-State Evolution}
\label{alg:Generation_State_Evolution}
\begin{algorithmic}[1]

\Require Context $\mathcal{X}'=(\mathcal{Q}, \mathcal{S}_{\text{sub}}, \mathcal{K}', \mathcal{K}_{\text{align}})$, LLM $\mathcal{M}$
\Ensure The final SQL query $\mathcal{Y}$

\State $\mathcal{H} \leftarrow \emptyset$
\State $t \leftarrow 0$
\State $\mathcal{Y}_1 \leftarrow \text{GenInitialQuery}(\mathcal{X}')$

\Loop 
    \State $t \leftarrow t + 1$
    \State $\mathcal{R}_t \leftarrow \text{Execute}(\mathcal{Y}_t)$
    \State $s_t \leftarrow (\mathcal{Y}_t, \mathcal{R}_t)$
    \State $\mathcal{H} \leftarrow \mathcal{H} \cup \{s_t\}$
    
    \State $\alpha \leftarrow \text{SelectAction}(\mathcal{X'}, \mathcal{H})$
    
    \If{$\alpha = \text{Finalize}$}
        \State $\mathcal{Y} \leftarrow \text{GenFinalQuery}(\mathcal{X'}, \mathcal{H})$
        \State \textbf{break}
    \Else
        \State $\mathcal{Y}_{t+1} \leftarrow \text{GenNextQuery}(\mathcal{H}, \alpha)$
    \EndIf
\EndLoop

\State \textbf{return} $\mathcal{Y}$

\end{algorithmic}
\end{algorithm}

\section{Compared Baselines}
\label{app:baseline}
For a fair and comprehensive comparison, we selected appropriate state-of-the-art baseline methods for each benchmark. 
On the \textbf{Spider 2.0-Snow dataset}~\cite{lei2024spider2}, we chose the representative agentic framework Spider-Agent~\cite{lei2024spider2} and ReFoRCE~\cite{deng2025reforcetexttosqlagentselfrefinement}, which is based on multi-path generation and voting, as our primary baselines. 
For the \textbf{BIRD dataset}~\cite{li2023llm}, our comparison encompasses four mainstream categories of methods: 
(1)~\textbf{Hybrid methods}, including CHASE-SQL~\cite{pourreza2024chasesqlmultipathreasoningpreference}, XiYan-SQL~\cite{liu2025xiyansqlnovelmultigeneratorframework}, the method from~\cite{shkapenyuk2025automaticmetadataextractiontexttosql}, and Contextual-SQL~\cite{agrawal2025text2sql}; 
(2)~\textbf{Post-training methods}, such as XiYanSQL-QwenCoder-32B~\cite{liu2025xiyansqlnovelmultigeneratorframework}, Arctic-Text2SQL-R1-7B~\cite{yao2025arctictext2sqlr1simplerewardsstrong}, Databricks-RLVR-32B~\cite{ali2025stateoftheartsqlreasoningmodel}, and OmniSQL-32B~\cite{li2025omnisqlsynthesizinghighqualitytexttosql}; 
(3)~\textbf{In-Context Learning}, including DAIL-SQL~\cite{dail_sql}, MCS-SQL~\cite{lee2024mcssqlleveragingmultipleprompts}, RSL-SQL~\cite{cao2024rsl}, and OpenSearch-SQL~\cite{xie2025opensearchsqlenhancingtexttosqldynamic}; 
and (4)~\textbf{Zero-shot methods}, such as DIN-SQL~\cite{pourreza2023dinsqldecomposedincontextlearning}, MAC-SQL~\cite{wang2025macsqlmultiagentcollaborativeframework}, and MAG-SQL~\cite{xie2024magsqlmultiagentgenerativeapproach}.
Furthermore, to ensure a fair comparison, we re-evaluated the performance of several baselines using their open-source implementations under a unified model and experimental environment.

\section{Error Analyses}

\label{app:error_ana}
In our evaluation on the Spider 2-Snow dataset, we analyzed 121 samples that have corresponding gold SQL queries. Of these, the 70 cases that DSR-SQL failed to predict correctly were selected for our manual analysis. Despite the limited sample size, we believe this analysis reveals generalizable error patterns. Unlike previous studies, our analysis introduces the concept of DSR-SQL's sub-states, enabling a more precise localization of error origins.

Our findings reveal that Text-to-SQL errors are often not attributable to a single cause but rather exhibit a cascading effect. For instance, the root cause of some Schema Linking errors can be traced back to an initial Intent Misunderstanding. Therefore, in analyzing each failure case, we documented and categorized all error types involved, rather than assigning a single, primary cause.
\begin{figure*}[htbp]
\centering
\includegraphics[width=1\linewidth]{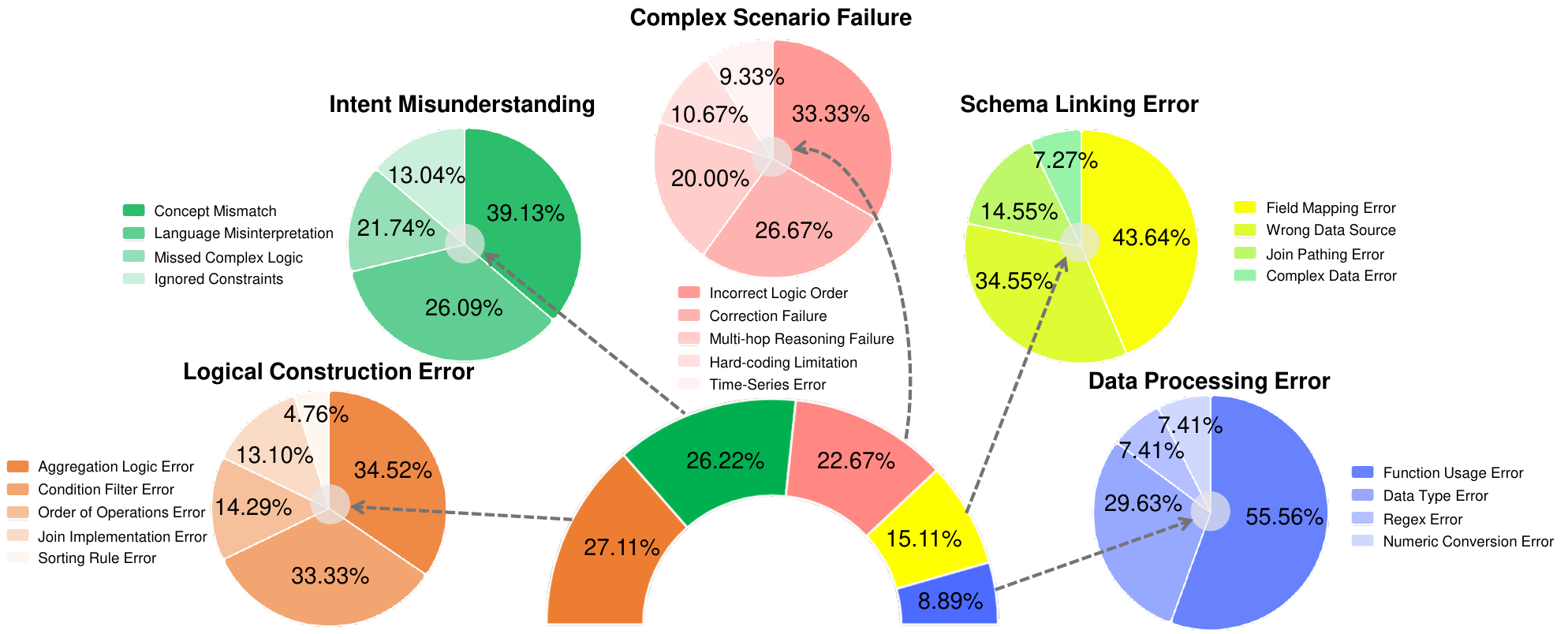}
\caption{Error Statistics of Spider2.0-Snow }
\label{fig:error_ana}
\end{figure*}

As shown in Figure~\ref{fig:error_ana}, we classify the errors into five major categories: \textbf{Logical Construction Error}, \textbf{Intent Misunderstanding}, \textbf{Complex Scenario Failure}, \textbf{Schema Linking Error}, and \textbf{Data Processing Error}. We elaborate on each category and its specific sub-types below:

\paragraph{Logical Construction Error}
This was the most prevalent category, accounting for approximately 27.11\% of errors. It refers to cases where the model partially understood the user's intent but deviated in constructing the SQL logic.
\begin{itemize}
    \item \textbf{Aggregation logic error:} The generated SQL failed to accurately reflect the user's aggregation intent when summarizing, counting, or grouping data. The core issue is a deviation in the data summarization method, leading to results with incorrect values, structure, or granularity.
    \item \textbf{Condition filter error:} The model correctly identified the filtering intent and attempted to construct conditions using \texttt{WHERE}, \texttt{HAVING}, or \texttt{JOIN ON}, but the resulting logic did not accurately or completely capture the user's constraints. This often stems from the model's unfamiliarity with database-specific vocabulary combined with imprecise user questions.
    \item \textbf{Order of operations error:} The model correctly identified most of the necessary columns and tables but erred in translating the mathematical logic or computational steps from natural language into SQL. The error lies not in selecting the wrong data, but in the incorrect method of calculation. For example, in a gene expression analysis, the predicted SQL calculated \texttt{LOG(AVG(value) + 1)}, which aligns with common intuition, whereas the domain convention is \texttt{AVG(LOG(value + 1))}. Due to the non-linearity of the logarithm function, these two orders yield vastly different results.
    \item \textbf{Join Implementation Error:} The generated SQL failed to accurately reflect the implicit or explicit data relationships required to combine data from multiple tables. This fundamentally corrupts the data foundation of the query, guaranteeing an incorrect final result even if subsequent logic is sound. Examples include selecting the wrong \texttt{JOIN} keyword (e.g., \texttt{INNER JOIN} vs. \texttt{LEFT JOIN}), leading to improper inclusion or exclusion of rows.
    \item \textbf{Sorting rule error:} The model recognized the need for sorting or ranking but failed to implement the logic according to the sorting criteria, direction, or priority explicitly or implicitly stated by the user.
\end{itemize}

\paragraph{Intent Misunderstanding}
This category was the root cause of many failures, accounting for approximately 26.22\% of errors. It typically manifests as the model's failure to correctly grasp the core intent, business logic, or key constraints of the user's question.
\begin{itemize}
    \item \textbf{Concept Mismatch:} The model identified a key business term or concept, but the \textbf{operational definition} or \textbf{computational scope} it used to translate it into SQL logic was inconsistent with the gold query's intent. For example, for the term "frequency," the LLM interpreted it as the physical count of a text's occurrences in a file, whereas the gold standard defined it as the total number of unique (repository, language) pairs associated with that text.
    \item \textbf{Language Misinterpretation:} Although the model parsed the keywords and syntax of the question, it failed to capture the deeper semantics, contextual constraints, or implicit logical relationships. This deviation is not a simple mapping error but arises during the translation from "intent" to "logic." For instance, translating "lacking a 'wikidata' tag" into a logically impossible condition such as \texttt{key = 'wikidata' AND key IS NULL}.
    \item \textbf{Missed Complex Logic:} When a user's question required multiple sequential and dependent computational steps, the model incorrectly simplified, merged, or flattened this complex process into a single step.
    \item \textbf{Ignored Constraints:} The model understood the core query objective but generated SQL that failed to adhere to all specified conditions, resulting in a dataset that was too broad, improperly filtered, or logically incomplete.
\end{itemize}

\paragraph{Complex Scenario Failure}
This category refers to failures in scenarios requiring multi-step reasoning or complex business logic.
\begin{itemize}
    \item \textbf{Incorrect Logic Order:} When a question required a specific sequence of logical steps (e.g., filtering, then aggregating, then calculating), the model failed to arrange these steps correctly. The error lies in the overall data flow and processing pipeline of the query, even if individual SQL clauses are syntactically correct.
    \item \textbf{Correction Failure:} When attempting to recover from an intermediate error (such as a query returning an empty set), the model not only failed to correct the logical flaw but also made a mistaken causal attribution (often concluding "the data does not exist"). Based on this false conclusion, it completely distorted the user's original intent, leading to an invalid or irrelevant final SQL query.
    \item \textbf{Multi-hop Reasoning Failure:} When addressing complex questions that require traversing one or more intermediate entities or logical steps to establish a data relationship, the model failed to construct the complete query logic. This reflects an inability to discover or correctly utilize non-direct, implicit connection paths between entities.
    \item \textbf{Hard-coding Limitation:} Within the DSR-SQL framework, the LLM, upon observing intermediate execution results, may opportunistically hard-code partial results from a previous step into subsequent queries. This practice undermines the query's generalizability.
    \item \textbf{Time-Series Error:} When processing time-dimensional data, the model failed to correctly understand and implement complex time-series analysis logic. This involved a fundamental misunderstanding of temporal continuity, dependencies, and window calculations, leading to incorrect time-series metrics.
\end{itemize}

\paragraph{Schema Linking Error}
This category refers to the model's failure to correctly understand the database structure and establish the correct mapping from natural language to database elements.
\begin{itemize}
    \item \textbf{Field Mapping Error:} When a table contained columns with highly similar names or functions, the model struggled to make the correct choice, especially when deeper semantic analysis was required for differentiation. For example, distinguishing between \texttt{assignee} and \texttt{assignee\_harmonized}, where the latter is a standardized field that provides more complete results.
    \item \textbf{Wrong Data Source:} When the database contained multiple tables with overlapping functions, similar names, but different data granularities or versions, the model failed to identify the most authoritative or best-matching source. This is one of the most common and impactful linking errors.
    \item \textbf{Join Pathing Error:} The model failed to correctly identify, select, or construct the inter-table join paths necessary to answer the user's question. This is a fundamental logical error, such as failing to find the correct "bridge table" or choosing an incorrect join path.
    \item \textbf{Complex Data Type Error:} The model was unable to generate correct SQL to query data stored in semi-structured or non-atomic columns (e.g., JSON, ARRAY). It tended to apply query logic suited for flat tables, failing when it needed to parse, un-nest, or navigate within complex structures.
\end{itemize}

\paragraph{Data Processing Error}
This category includes technical errors made by the model during data processing and transformation.
\begin{itemize}
    \item \textbf{Function Usage Error:} The generated SQL might be syntactically valid but failed to accurately implement the user's business logic due to a misunderstanding of the selection, parameters, or application context of an SQL function. For example, while the \texttt{TO\_TIMESTAMP\_*} family of functions all convert timestamps, the model might confuse their subtle differences in time zone handling.
    \item \textbf{Data Type Error:} Although the model identified a column's basic type (e.g., number, string), it erred in handling its specific format, intrinsic semantics, or operational requirements, leading to format misunderstandings or improper type conversions.
    \item \textbf{Regex Error:} The regular expression used in the SQL was unable to accurately, completely, or robustly match or extract the required information from the target text field. This usually stemmed from an insufficient understanding of the target text's structure, diversity, or edge cases.
    \item \textbf{Numeric Conversion Error:} When performing mathematical operations or type conversions on numerical values (to change their unit, scale, or precision), the process did not align with the user's requirements or the data's intrinsic logic, leading to skewed numerical results.
\end{itemize}

Overall, our analysis demonstrates that even powerful reasoning models still face numerous challenges when tackling highly difficult Text-to-SQL benchmarks like Spider 2-Snow. Looking ahead, future research in the Text-to-SQL domain should continue to focus on several key areas: more refined task context management, more accurate alignment of domain-specific database knowledge with user intent, and enhancing model adaptability to different SQL dialects.

\section{Case Study: Leveraging 'Revise' Actions for Feedback-Driven Correction}
\begin{figure*}[t]
\centering
\includegraphics[width=1\linewidth]{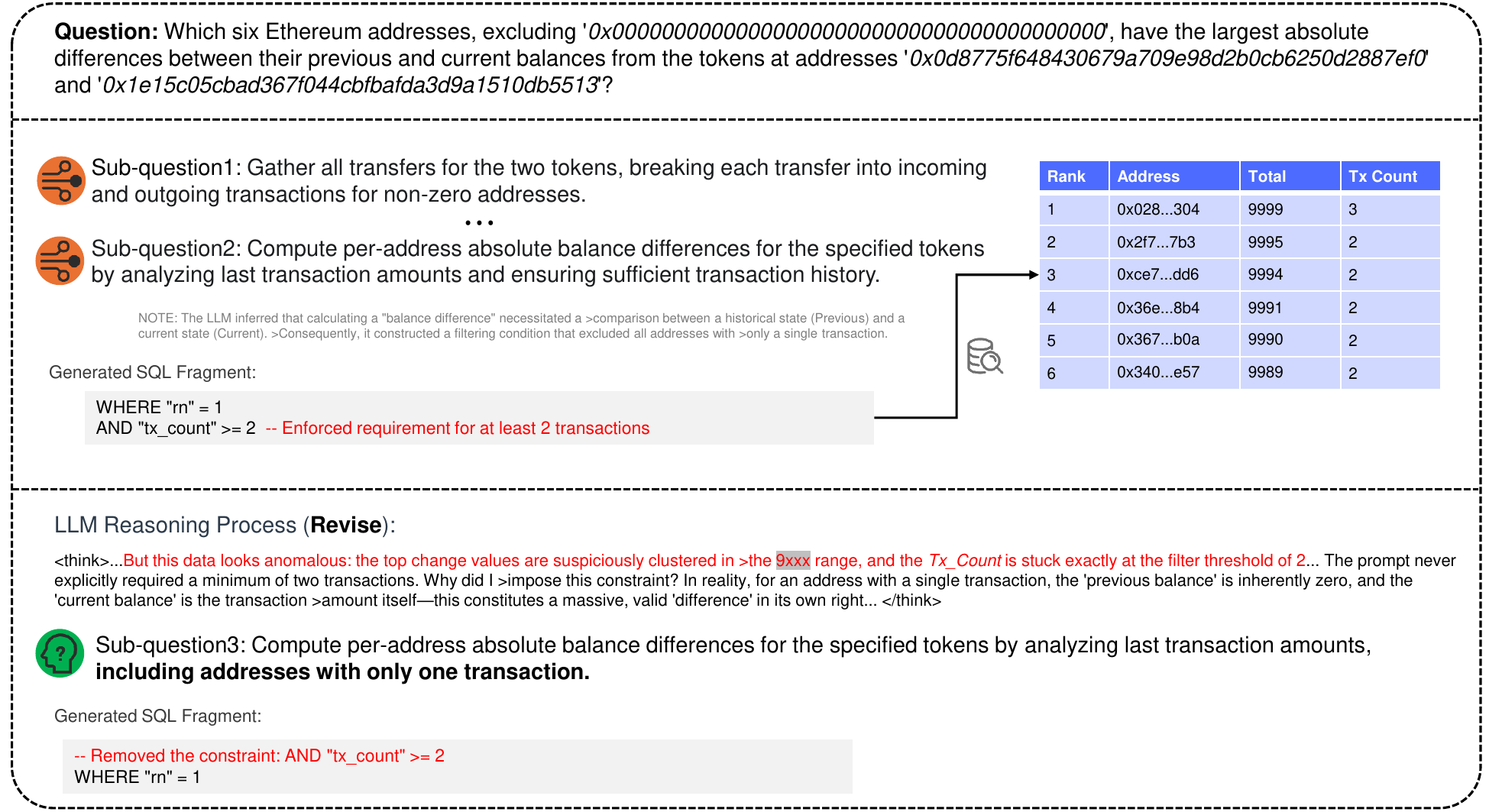}
\caption{Leveraging 'Revise' Actions for Feedback-Driven Correction }
\label{fig:Case_study_Revise}
\end{figure*}

We present a case study that demonstrates the LLM's ability to successfully leverage execution feedback to correct its SQL generation trajectory, a process we term the \textbf{Refinement Path}. As shown in Figure~\ref{fig:Case_study_Revise}, upon observing the execution feedback from the preceding sub-SQL, the LLM re-evaluated the current business logic and implemented a corresponding revision step. It expanded the calculation scope from 'Active Accounts' to 'All Accounts,' ensuring that single-transaction surges (e.g., spiking from 0 to 10 million) were correctly identified as the 'largest differences.'

\section{Case Study: How Execution Feedback Drives Problem Solving}

\begin{figure*}[t]
\centering
\includegraphics[width=1\linewidth]{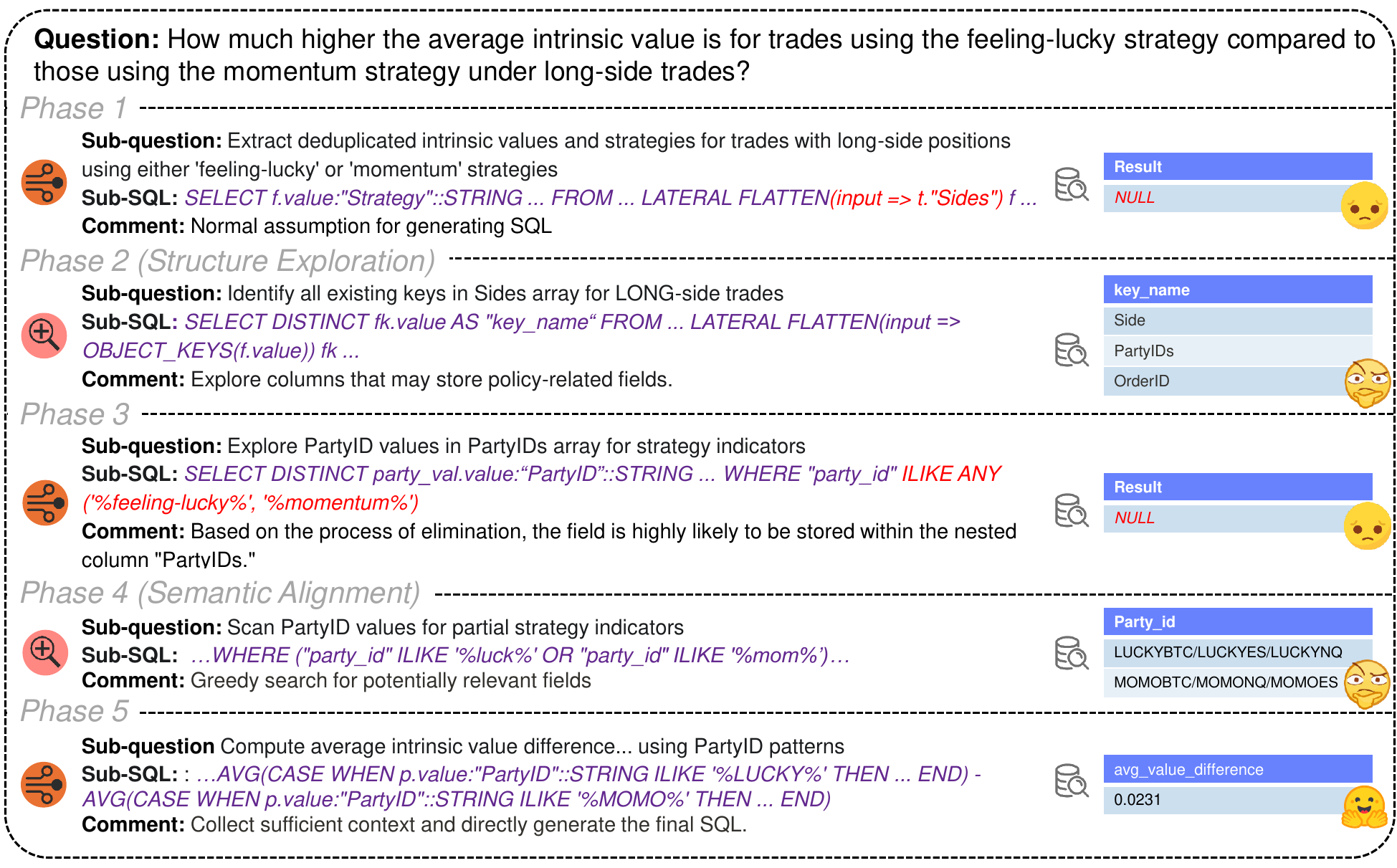}
\caption{How Execution Feedback Drives Problem Solving}
\label{fig:Case_study_EF}
\end{figure*}

The utilization of execution feedback is a core tenet of the DSR-SQL framework, and we present a successful case study in Figure~\ref{fig:Case_study_EF}. As illustrated, in the initial stage, the LLM relies on its general knowledge to analyze the user's question and generate a corresponding SQL query, which results in an empty execution result. In Stage 2, the LLM initiates an \textbf{Explore} action to identify the column where the 'strategy' entity is likely stored. Stage 4 is the most pivotal phase: through a greedy search, the LLM discovers the association between the user-expressed entity and its corresponding representation in the database, thereby achieving \textbf{Semantic Alignment}. At this point, having gathered all necessary context to answer the question, the model proceeds to the final generation stage. This case clearly demonstrates how DSR-SQL leverages execution feedback to familiarize itself with the structure of an unknown database, progressively aligning user intent. It showcases the practical utility of DSR-SQL in navigating unfamiliar and complex scenarios.

\section{Other results}
Subsequent to the submission of our manuscript, the official evaluation script for Spider 2.0 underwent a major revision, which may lead to inconsistencies in reported scores. Although all experiments in our paper were conducted using the original version of the script, we report comparison results on both the old and new versions in Table~\ref{tab:spider_new_compact} to ensure transparency and consistency. Furthermore, we also report the performance of DSR-SQL on the multi-dialect Spider 2.0-Lite benchmark, on which we achieved a third-place ranking at the time of our submission.

\begin{table}[h]
    \centering
    \begin{tabular}{lcc}
        \toprule
        Benchmark & Script Version & EX (\%) \\
        \midrule
        \multirow{2}{*}{Spider 2.0-Snow} & Old & 35.28 \\
                                         & New & 52.83 \\
        \midrule
        \multirow{2}{*}{Spider 2.0-Lite} & Old & - \\
                                         & New & 46.80 \\
        \bottomrule
    \end{tabular}
    \caption{Performance of DSR-SQL on the Spider 2.0 benchmark, along with other results.}
    \label{tab:spider_new_compact}
\end{table}

\section{Disclosure of Large Language Model Usage}
During the development phase, a LLM was utilized to assist in the generation of code for several utility functions. Additionally, an LLM was employed during the manuscript preparation phase for grammatical polishing and refinement. All authors have reviewed and verified the correctness of the LLM-generated content and unanimously consent to its inclusion in this work.

\end{document}